\documentclass[letterpaper, 10 pt, conference,dvipsnames]{ieeeconf}  

\IEEEoverridecommandlockouts                              
\usepackage[utf8]{inputenc}
\usepackage[T1]{fontenc}

\usepackage{color}
\usepackage{xcolor}
\usepackage{commath}

\usepackage{graphicx} 
\usepackage{amsmath} 
\usepackage{amssymb}  
\usepackage{cancel}
\usepackage{verbatim}
\usepackage{algorithm}
\usepackage{algpseudocode}
\usepackage{float}
\usepackage{tikz}
\usepackage{hyperref}
\usepackage{comment}
\usepackage{mathtools}
\usepackage{tabularx,tabulary}
\usetikzlibrary{automata,positioning}
\tikzstyle{node}=[align=center]

\let\labelindent\relax
\usepackage[inline]{enumitem}

\usepackage{comment}
\usepackage{xcolor}
\usepackage{xspace}
\usepackage[normalem]{ulem}

\definecolor{OliveGreen}{RGB}{0,200,25}
\newcommand{\red}[1]{\textcolor{red}{#1}}

\newcommand{\darkgreen}[1]{\textcolor{OliveGreen}{#1}}

\newcommand{\ie}{i.\,e.}
\newcommand{\eg}{e.\,g.}

\newcommand{\replaced}[2]{\red{\ifmmode\text{\sout{\ensuremath{#1}}}\else\sout{#1}\fi}\darkgreen{#2}}
\newcommand{\removed}[1]{\red{\ifmmode\text{\sout{\ensuremath{#1}}}\else\sout{#1}\fi}}

	\renewcommand{\replaced}[2]{#2}
	\renewcommand{\removed}[1]{}
	
\newcommand{\removedfootnote}[1]{\footnote{\removed{#1}}}
\newcommand{\removedsubsection}[1]{\subsection{\texorpdfstring{\removed{#1}}{#1}}}

	\renewcommand{\removedfootnote}[1]{}
	\renewcommand{\removedsubsection}[1]{}
	
	\renewcommand{\removedsubsection}[1]{}

\title{\LARGE \bf
Safety Evaluation of Robot Systems via Uncertainty Quantification
}

\author{Woo-Jeong Baek and Torsten Kröger
\thanks{The authors are with the Institute for Anthropomatics and Robotics (IAR-IPR), Karlsruhe Institute of Technology (KIT)
        {\tt\small \{baek, torsten\}@kit.edu}.}
\thanks{The authors like to thank Tamim Asfour for his guidance and Lars Berscheid for fruitful discussions throughout this work.}
}

\begin{document}
\graphicspath{ {./figures/} }

\maketitle
\thispagestyle{empty}
\pagestyle{empty}

\begin{abstract}
In this paper, we present an approach for quantifying the propagated uncertainty of robot systems in an online and data-driven manner. 
Especially in Human-Robot Collaboration, keeping track of the safety compliance during run time is essential: Misclassifying dangerous situations as safe might result in severe accidents. 
According to official regulations (\eg, ISO standards), safety in industrial robot applications depends on critical parameters, such as the distance and relative velocity between humans and robots. 
However, safety can only be assured given a measure for the reliability of these parameters. 
While different risk detection and mitigation approaches exist in literature, a measure that can be used to evaluate safety limits online, and succinctly implies whether a situation is safe or dangerous, is missing to date. 
Motivated by this, we introduce a generalizable method for calculating the \emph{propagated measurement uncertainty} of arbitrary parameters, that captures the accumulated uncertainty originating from sensory devices and environmental disturbances of the system. 
To show that our approach delivers correct results, we perform validation experiments in simulation. 
In addition, we employ our method in two real-world settings and demonstrate how quantifying the propagated uncertainty of critical parameters facilitates assessing safety online in Human-Robot Collaboration. 
\end{abstract}

\section{INTRODUCTION}
Within the realm of robotics, safety has attracted great interest in recent years. 
Especially in human-involved systems, providing a safe environment throughout operation time is inevitably important. 
In current industrial robot applications, robots are clearly separated from the human workers or allowed to only perform very slow movements to guarantee human safety. 
To enable flexible and more efficient collaboration scenarios, where human and robot can share their workspace, an online safety evaluation technique is necessary. 
According to official standards provided by the International Organization for Standardization (ISO), safety limits in Human-Robot Collaboration (HRC) are defined via critical parameters as the distance or relative velocity between humans and robots. 
In fact, a situation is defined as \emph{safe} as long as specified parameter thresholds are kept. 
 \begin{figure}[ht!!]
      \centering
   \includegraphics[scale=0.28]{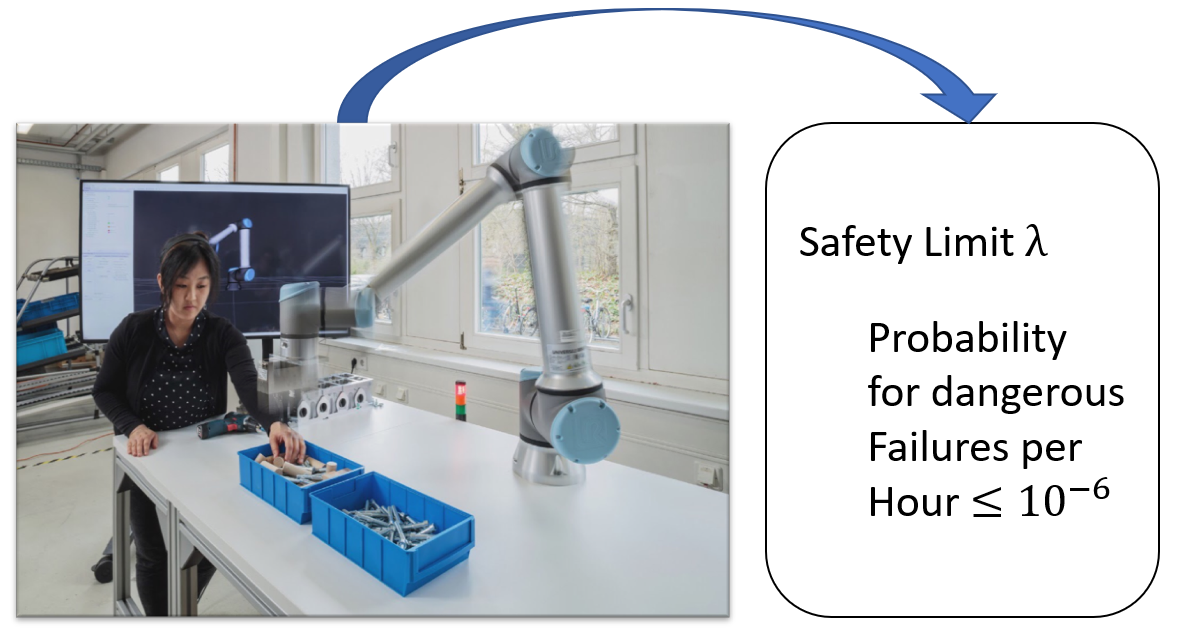}
     \caption{We present a generalizable method for quantifying the propagated measurement uncertainty (PMU) to evaluate safety limits in robot systems. According to official regulations, we refer to the probability for dangerous failures per hour (PFDH) to distinguish safe from dangerous situations.}
      \label{fig_overview}
   \end{figure}
One challenge we face in dynamic environments is that these critical parameters change continuously (\eg, moving humans and robots). 
On top of that, the strict requirement in standard ISO 13849 to keep the rate for the occurrence of dangerous failures below $10^{-6}$ per hour, equaling one tolerated accident in approximately 114 years, amplifies the burden on assuring safety in HRC. 
In this contribution, we provide one step toward the development of safe robot systems by arguing that evaluating safety requires quantifying measurement uncertainties of the critical parameters. 
In particular, we state that the \emph{reliability} of the parameter estimation plays a significant role and that it can be captured by the \emph{propagated measurement uncertainty (PMU)}. 
To be specific, the PMU reflects how much a parameter might deviate from the measured value. 
For instance, a PMU of $0.2\,m$ for an estimated distance of $1.5\,m$ between human and robot would mean that the actual distance could be lower or higher by $0.2\,m$ with a certain probability.
While overestimating the distance between humans and robots might result in neglecting dangerous situations and finally to severe accidents, an underestimation would initiate false alarms and affect the efficiency. 
However, an online safety evaluation method that accounts for the reliability of the critical parameters does not exist to date. 
To fill this gap, we introduce a generalizable method to quantify the PMU: a scalar value reflecting the total amount of uncertainties accumulated along the system pipeline. 
Particularly, our approach accounts for technical limitations of system components and environmental disturbances (\eg, lightning conditions). 
We refer to our previous work \cite{Baek2022} to determine the uncertainties of components, where no manufacturer specifications are available. 
Furthermore, we employ a Monte-Carlo sampling based technique suggested in the Guide to the Expression of Uncertainties in Measurements (GUM) \cite{GUM_S2008} to propagate uncertainties originating from different sources. 
In addition, we present how the PMU can be interpreted in the context of safety limits by referring to the threshold defined in standard ISO 13849. 
After validating the correctness of our method in a PyBullet simulation environment, we highlight the applicability in real-world scenarios by demonstrating how the online PMU calculation serves as a method for evaluating safety limits in HRC. 

\begin{figure*}[htbp]
      \centering
   \includegraphics[width=1.0\textwidth]{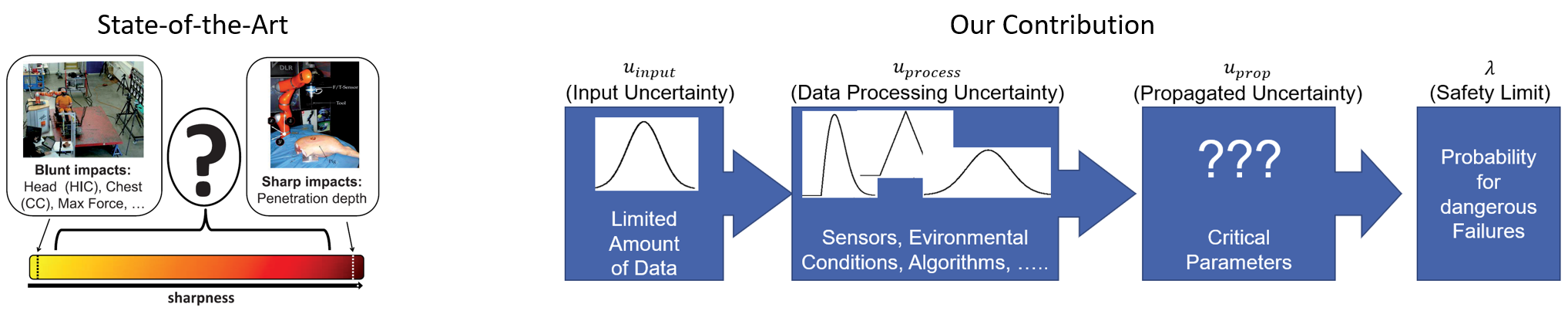}
     \caption{Left: The contribution in \cite{Haddadin2012} presents a control framework based on injury analyses on pig samples. In doing so, a safety evaluation method accounting for the severity of injuries is developed. Right: We present a generalizable, data-driven approach to quantify the propagated uncertainty on critical parameters. Furthermore, we show how it can me mapped on an arbitrary safety limit that is characterized by the probability for dangerous failures.}
      \label{fig:sota_comp}
   \end{figure*}

\section{Related Work} \label{sota}
Our contribution can be assigned to two research areas: 
Regarding the safety evaluation, our approach belongs to safety assessment methods in HRC. 
On the other hand, uncertainty calculation in robot systems has been studied extensively in subdomains as robot navigation. 
In the following, we outline contributions from both areas. 

\subsection{Safety Evaluation in Human-Robot Collaboration (HRC)}
Existing methods for safety evaluation can be largely categorized as follows:
The first category covers simulation based risk assessment methods which are conducted offline. 
For example in \cite{Wadekar2018} and \cite{Inam2018}, the human-robot work spaces are modeled in simulation environments. 
Wadekar et al. suggest to provide a list of tasks which are executed by humans and robots in \cite{Wadekar2018}. 
Based on this list, possible dangerous situations are identified by referring to ISO standard requirements. 
In doing so, this work proposes a method that acts as a basis for designing safe collaborative work cells. 
In \cite{Inam2018}, the authors first specify human-robot interaction scenarios and conduct a risk assessment by means of Hazard Operability (HAZOP) \cite{Guiochet13}. 
To reduce the occurrence of risks, Inam et al. propose a method based on scene graphs identified in the simulation environment V-REP. 
Similarly, Askarpour et al. introduce a safety analysis methodology SAFER-HRC in \cite{Askarpour2016} by means of an Operator-Robot-Layout (ORL) model.
This model captures the dynamics of human-robot interaction. 
For risk detection purposes, a formal logic language is presented: After generating a list that specifies the tasks between human and robot, this list is converted to formal models to estimate the severity of possible accidents. 
To evaluate safety, an iterative analysis is conducted to capture dangerous situations. 
While these contributions present promising approaches for risk identification in simulation, one of the main drawbacks is given by the sim-to-real gap that limits the transfer of detected risks to the real world. 
The second line of works focuses on the development of control strategies in HRC. 
For instance, the safety evaluation is inherently incorporated in the control architecture by Zanchettin et al. in \cite{Zanchettin2016}. 
Based on pre-defined safety constraints, sensor signals are interpreted by the controller to select safe actions for the robot. 
While worst case uncertainties are considered by a fixed parameter, a thorough characterization regarding their amount or behavior is not addressed. 
One further interesting work is presented by Haddadin et al. in \cite{Haddadin2012}. 
With the goal to account for the severity of collisions, the authors perform skin injury analyses on pig samples. 
By classifying the severity of injuries for various impact masses, the functional relationship with the robot velocity is determined. 
These findings are adopted in the control architecture to provide a motion supervising component that avoids collisions in real-time. 
The third line of works addresses the development and incorporation of speed and separation monitoring (SSM) methods. 
The purpose of the SSM mode suggested in standard ISO 10218 \cite{ISO10218} is to maintain a certain speed and distance between robots and humans. 
In \cite{Lacevic2010} and \cite{Lacevic2011}, Lacevic et al. introduce so-called kinetostatic danger fields for hazard identification. 
These fields capture possible collision objects and calculate the velocity vector pointing from the robot to these objects, thus allowing to deduce the actual danger level. 
In a further contribution \cite{Polverini2014}, this approach is extended and generalized to dynamic scenarios. 
A more thorough discussion of the formalization of SSM is provided in \cite{Marvel2017}. 
Here, Marvel et al. elaborate on the implementation of SSM in collaborative work cells and point out the necessity of integrating uncertainties in the safety evaluation pipeline. 
In fact, Marvel et al. focus on evaluating the SSM criterion and suggest three metrics: the severity of possible collisions, the separation distance between robot and potential obstacles and the measurement uncertainty. 
In contrast, Lasota et al. present the development of a real-time safety system that controls the robot speed and position based on human tracking in \cite{Lasota2014a}. 
Making use of the PhaseSpace motion capture system, the uncertainty of the tracking process is assumed to be negligible. 
Based on this assumption, a control algorithm that incorporates the distance between human and robot is developed. 

\subsection{Uncertainty Propagation in Robotics}
In this work, we take the metrological viewpoint of uncertainties as elaborated in \cite{GUM2008} and \cite{GUM_S2008}. 
Our goal is to provide a measure that reflects the reliability of critical parameters. 
Although propagating uncertainties has not been addressed in the context of safety evaluation in HRC to date, works from the field of mobile robot navigation show similarities from methodological perspective. 
For example, the contributions in \cite{Luo2020} and \cite{Arvanitakis2017} suggest probabilistic approaches to capture uncertainties. 
In particular, the authors in \cite{Luo2020} define probabilistic safety barrier certificates based on Gaussian distributions. 
These certificates define the control space and guarantee that a specified safety condition is held for each time step. 
On the other hand, Arvanitakis et al. directly refer to measurement uncertainties and aim to alleviate the occurrence of inaccurate navigation behavior by introducing an uncertainty space in \cite{Arvanitakis2017}.
While an effective control law is derived, it is assumed that the uncertainty space reflecting its amount and behavior is known beforehand. 
An overview is provided by Thrun et al. in \cite{Thrun2006}, that summarizes a variety of uncertainty estimation techniques. 
The contributions of Roy et al. in \cite{Roy1999} and Giancola et al. in \cite{Giancola2018} show the highest similarities to our work:
The first paper presents a method that uses the Bayes formulation to infer external sensor data with the robot position. 
In particular, uncertainties of all sensor components are taken into consideration to perform a Markov localization. 
Based on the results, the authors introduce a certainty parameter that allows to reversely quantify the uncertainty. 
In contrast, Giancola et al. study the uncertainty behavior of three camera types (Time of Flight, Structured Light and Active Stereo) in the face of environmental disturbances, thereby discussing the influence of parameters. 
The survey in \cite{Hill2020} provides an overview of three uncertainty propagation techniques (Kalman filter, particle filter and a calculus based method) and point out the strengths and weaknesses.  
Apart from these works, there exist several contributions as \cite{Riaz2020}, \cite{Loquercio2020} that use learning methods to account for uncertainties. 
However, while alleviating undesired consequences due to uncertainties might be possible, a thorough characterization cannot be conducted due to the black-box character of neural networks.

\section{Measurement Uncertainties and Safety} \label{sec:meas_safety}
The goal of this paper lies in developing an online safety evaluation method that provides the PMU accompanied by the information whether a situation is safe or not. 
According to official regulations as standard ISO 12100 \cite{ISO12100}, the term \textit{risk} of an incident $i$ is defined by its probability of occurrence $Pr(i)$ multiplied by the respective severity $s(i)$: 
\begin{equation}
    risk (i) = Pr(i)\cdot s(i).
    \label{eq:risk}
\end{equation}
Here, the severity reflects the criticality of consequences: In HRC, the severity is measured by the level of injuries on the human body. 
The underlying motivation of our contribution is that a perfectly accurate perception of the environment (\eg, lightning conditions, temperature fluctuations) and all relevant data (\eg, robot state and human state) for all time steps would yield a probability of $Pr(i)=0\,\%$. 
In other words, given real-time capability and a 100\% certainty for all happenings, we would be able to predict and thus avoid all risks before they occur. 
However, this cannot be achieved in the real world due to the limited perception capability and the missing knowledge on the system parameters as well as their relationships to each other.  
Nevertheless, an appropriate candidate that allows to capture the extent of undesired deviations caused by these shortcomings is given by the PMU that reflects how much a quantity might deviate from a measured value. 
For example, the human position in HRC usually underlies deviations due to the limited resolution of employed cameras, that may be further affected by changes in the lightning conditions. 
In the following, we propose a generalizable method to quantifying the PMU, thereby providing a data-driven method to evaluating safety limits online.  
To distinguish safe from unsafe situations, we refer to the \emph{probability for dangerous failures per hour (PFDH)} that is introduced by official ISO regulations.  

\section{Method}
First, we must specify for which parameter the uncertainty propagation shall be performed.
We denote this parameter with the \emph{attribute a} and the respective uncertainty with $u_a$. 
Often, the attribute $a$ is the output of a system, that is, the product data, that is captured and subsequently processed by technical tools. 
Hence, to determine the uncertainty $u_a$, the uncertainties from all system components that contribute to this output must be taken into consideration. 
According to \cite{GUM2008}, this requires a model for the functional relationship $f$ of the $n$ input parameters $x_1,...x_n$ given by 
\begin{equation}
  a \coloneqq   f(x_1,...,x_n),
  \label{eq:model_a}
\end{equation}
that describes how the attribute $a$ stands in relation to the input parameters $x_1,...x_n$ provided by the $n$ system components. 
For example, if we are interested in determining the position of an object, its state space coordinates would be possible candidates for the respective attribute. 
To calculate $u_a$, we must account for the uncertainties of all input parameters $x_i$ according to Eq.~\eqref{eq:model_a}. 

\subsection{Uncertainty Classification}
\label{subsec_uncclass}
In order to correctly perform the uncertainty propagation, each uncertainty source must be classified as follows. 
We distinguish two uncertainty types in analogy to \cite{GUM2008}:
\begin{description}
    \item Type A Uncertainties: These cover uncertainty sources like sensory devices, where the uncertainty is provided by a model (\eg, simulation) or the manufacturer. Throughout this work, we rely on the correctness of the information stated in data sheets or models. 
    \item Type B Uncertainties: In this case, no information of the uncertainty source is available. This means that the amount and behavior of Type B uncertainties must be estimated by measurements. One possibility is to leve\-rage the knowledge of conservation measures, that capture non-changing system characteristics. For instance, to estimate the uncertainty of an output signal $o$, that is known to stay constant over time, the conservation equation would be given by $\frac{\partial o}{\partial t} = 0.$
    The occurrence of any deviation from this equation is considered to draw from the uncertainty of the component that captures and processes the data to produce $o$. We refer to \cite{Baek2022} for details.  
\end{description}

 \begin{figure}[htbp]
      \centering
   \includegraphics[width=0.4\textwidth]{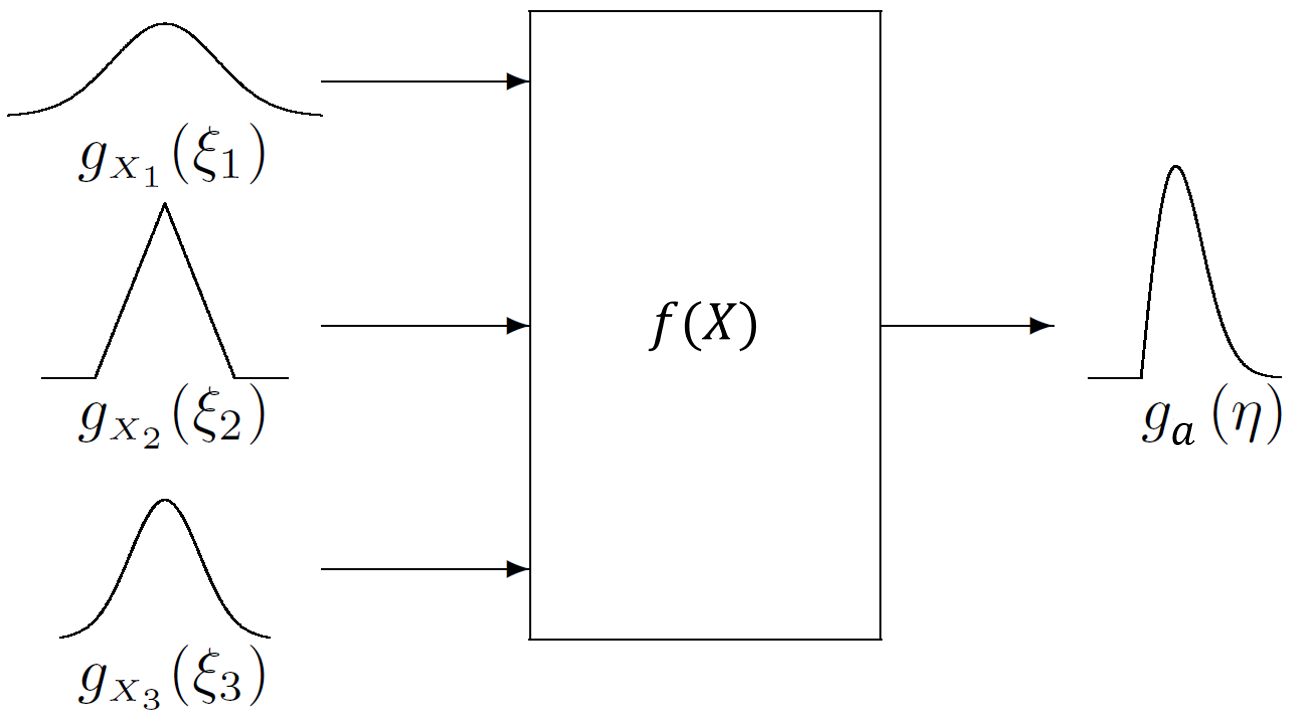}
     \caption{Illustration of the propagation of uncertainty distributions for $n=3$ independent input parameters $x_i$ adapted from \cite{GUM_S2008}. In contrast to cases, where the uncertainties of the input parameters are represented by scalar values, Monte-Carlo sampling must be applied to obtain the propagated uncertainty behavior $g_a$ of the attribute $a$.}
      \label{fig_unc_dist}
   \end{figure}
   
\subsection{Uncertainty Propagation}
\label{subsec_uncprop}
We now turn towards the calculation of the PMU of the attribute $a$. 
For negligible correlations between the input parameters $x_i$, the PMU denoted with $u_{prop}(a)$ can be determined via the local derivatives
\begin{equation}
    u_{prop}(a) = \sqrt{\sum_{i=1}^N \left(\frac{\partial a}{\partial x_i}\right)^2 \cdot u_{x_i}^2(a)}
    \label{eq:prop_uncorr}
\end{equation}
with the input uncertainties $u_{x_i}$ according to \cite{GUM2008}.
However, for correlated input parameters, Eq.~\eqref{eq:prop_uncorr} changes to
\begin{equation}
    u_{prop}(a) = \sqrt{\sum_{i=1}^N \sum_{j=1}^N \frac{\partial a}{\partial x_i}\frac{\partial a}{\partial x_j} u_{x_i, x_j}(a)},
    \label{eq:prop_corr}
\end{equation}
where $u_{x_i, x_j}(a)$ denotes the covariance between $x_i$ and $x_j$. 
However, these equations are only valid for small input uncertainties $u_{x_i}$, that is, for cases where their exact functional model can be neglected.  
We argue that a thorough safety evaluation according to Eq.~\eqref{eq:risk} requires to consider for the functional behavior of $u_{x_i}$. 
Thus, we suggest to replace the scalar value with 
$
    x_i \propto \mathcal{N}(\mu_i, \sigma_i),
$
where $\mu_i$ represents the measured value of $x_i$ and $\sigma_i$ the corresponding uncertainty.  
Since above equations Eq.~\eqref{eq:prop_uncorr}, Eq.~\eqref{eq:prop_corr} are defined for scalar uncertainties, we cannot directly make use of them to obtain $u_{prop}(a)$ anymore. 
Instead, we must propagate distributions according to the model equation Eq.~\eqref{eq:model_a}. 
To do so, we generate Monte Carlo (MC) samples and obtain the PMU $u_{prop}(a)$ with following steps (see \cite{GUM_S2008} for details): 
\begin{enumerate}
    \item Set number M of MC trials. 
    \item Create M vectors from the $n$ probability density functions assigned to input parameters $x_i$. 
    \item Evaluate value for attribute $a$ according to the model function $f$ in Eq.~\eqref{eq:model_a} for each of the M vectors.  
    \item Sort the M values of $a$ into increasing order to obtain a distribution function $G$ (see \cite{GUM_S2008}, Chapter 5.3 for details).
    \item Use this distribution function $G$ to approximate $f$ and the PMU $u_{prop}(a)$. 
\end{enumerate}

\begin{figure*}[ht!!!]
      \centering
   \includegraphics[width=1.0\textwidth]{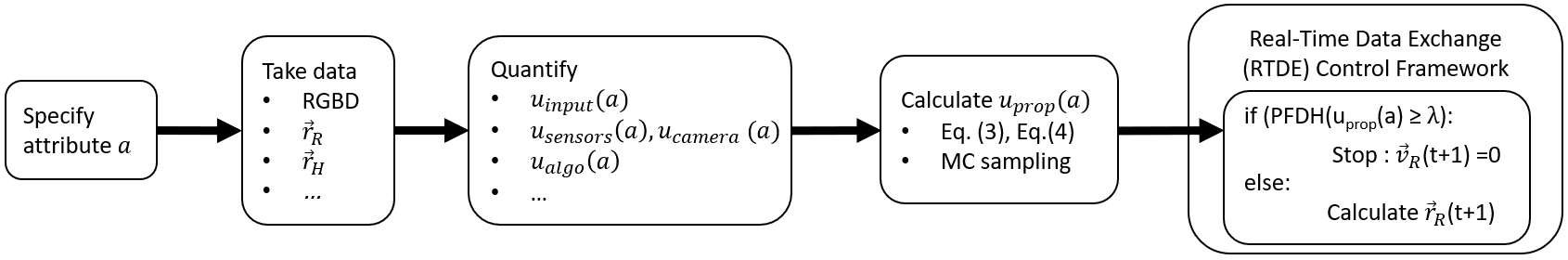}
     \caption{Based on the specified attribute $a$, we quantify the input uncertainties and the uncertainties originating from the data processing. In the next step, we calculate the PMU $u_{prop}(a)$ according to Eq.~\eqref{eq:prop_uncorr}, Eq.~\eqref{eq:prop_corr} and the MC sampling. By incorporating the mapping on the safety limit with Eq.~\eqref{eq:safety_lambda} into the RTDE control framework, the robot movement is stopped once the safety requirement is violated.}
      \label{fig:approach}
   \end{figure*}

\section{Mapping the PMU on the Safety Limit $\lambda$}
\label{sec_PMULambda}
In this Section, we specify how the PMU of critical parameters can be mapped on safety limits. 
As elaborated in Section \ref{sec:meas_safety}, two critical parameters in HRC are given by the distance $d$ and the relative velocity $v$ between humans and robots. 
Therefore, these parameters are treated as the attributes, \ie, 
\begin{equation}
    a \mapsto d;  \, \, \, \, a \mapsto v. 
    \label{eq:attribute_distance}
\end{equation}
By mapping the PMU of these attributes on the probability for dangerous failures per hour (PFDH), we accomplish the safety evaluation. 
While the PFDH is a measure introduced by the International Organization for Standardization (ISO), we stress that following steps are adaptable for arbitrary failure rates. 
We view the PMU as the possible deviation accumulated over all previous time steps. 
Also, a situation is considered as unsafe once the following condition is violated: 
\begin{equation}
a \pm u_a \leq \lambda_a,
\label{eq:safety_lambda}
\end{equation}
where $\lambda_a$ is the safety limit with respect to attribute $a$. 
For instance, these can be distance or velocity limits provided by ISO 10218: A standard defining explicit thresholds for the tolerated distance, velocity and force between humans and robots in speed and separation monitoring (SSM). 
In addition, the maximum failure rate of $\rm{PFDH} = 10^{-6}/h$ must be held according to ISO 13849 \cite{ISO13849}. 
For a known frame rate $f_p$ in object or human tracking, we obtain the PFDH for an attribute $a$ via the following expression: 
\begin{equation}
\mathrm{PFDH}(a) = N_d \cdot f_p \cdot 3600,
\label{eq:mapping}
\end{equation}
where $N_d$ is the number of incidents that violate Eq.~\eqref{eq:safety_lambda}. 

\section{Experiments}
\label{sec:exp}
We first verify the correctness of our method to highlight how the PMU calculation can be employed for safety evaluation. 

\subsection{Validation in PyBullet}
In contrast to technical tools, where the uncertainties are stated in the data sheets, the PMU of a real-world system is not available: 
Apart from the missing knowledge on how the parameters are related to each other, it is also not clear which parameters must be observed. 
For example, when tracking an object in real-world, we cannot guarantee with $100\,\%$ certainty, that the tracking is independent from the room temperature or other parameters. 
Thus, to validate our method, an environment that allows to control all parameters is required. 
To this end, we carry out proof-of-concept experiments in a PyBullet simulation environment. 
Specifically, we focus on tracking the 2D position of a UR10e robot end-effector by using a camera with a frame rate of $f_s=30\,\mathrm{fps}$. 
To detect the end-effector position, we apply a red mask on it. 
After calibrating the camera, we measure the center of the red mask based on the recorded data to obtain the end-effector coordinates. 
Here, the attribute $a$ and its uncertainty $u_a$ are given by  
$
    a \mapsto \Vec{r}_{s}(t)=(x_{s}, y_{s}); $ and $ u_a \mapsto u_{\Vec{r}_s}(t),
$
where $t$ stands for the time and $x_s$, $y_s$ denote the $x$, $y$-coordinates of the end-effector in Euclidean space, respectively. 
We record 6000 frames for each robot velocity settings, where $v_1 < v_2 < v_3$. 
We consider Gaussian-distributed uncertainties for the camera and the robot velocity. 
Specifically, we account for a resolution uncertainty of $0.02\,\%$ for the camera and an uncertainty of $10\,\%$ for the robot velocity. 
We set $M=10^5$ for the number of MC trials, considering that the camera uncertainty is correlated to the robot velocity uncertainty. 

\subsection{2D Marker Tracking under Real-World Conditions}
The goal of this experiment is to demonstrate the real-world applicability of our method. 
Again, we track the position of the UR10e robot end-effector by using the intel RealSense D435. 
By using the Real-Time Data Exchange (RTDE) interface provided by Universal Robots, we generate a point to point trajectory for the robot in horizontal direction.
The ground truth value of the end-effector joint is easily accessible by RTDE. 
We attach an ARUCO marker \cite{Garrido2014} on the robot end-effector to perform a 2D position tracking. 
As in the previous case, the tracking process is subject to uncertainties due to the robot velocity and the limited performance of the camera. 
After completing the calibration of the camera intel RealSense D435 into the world frame, we compare the actual end-effector position with the coordinates of the ARUCO marker determined by the RGB data. This allows us to verify whether the ground truth value lies in the estimated range including the uncertainty, \ie 
\begin{equation}
    \Vec{r}_A(t) \pm u_A(t) \geq \Vec{r}_{GT}. 
    \label{eq:discrepancy}
\end{equation}
Here, $\Vec{r}_A(t)$ stands for the end-effector position obtained by tracking the ARUCO marker, $u_A(t)$ the corresponding PMU and $\Vec{r}_{GT}$ the end-effector ground truth position provided by RTDE. 
We perform experiments for three robot velocity settings $v_1 < v_2 < v_3$. 
For each setting, we collect 900 frames and set the number of MC trials to $M=10^5$. 
To obtain the ground truth uncertainty, we refer to the data sheet of the intel RealSense D435 devices stating an uncertainty of $0.02\,\%$. 

\begin{figure*}[htbp]
      \centering
   \includegraphics[scale=0.35]{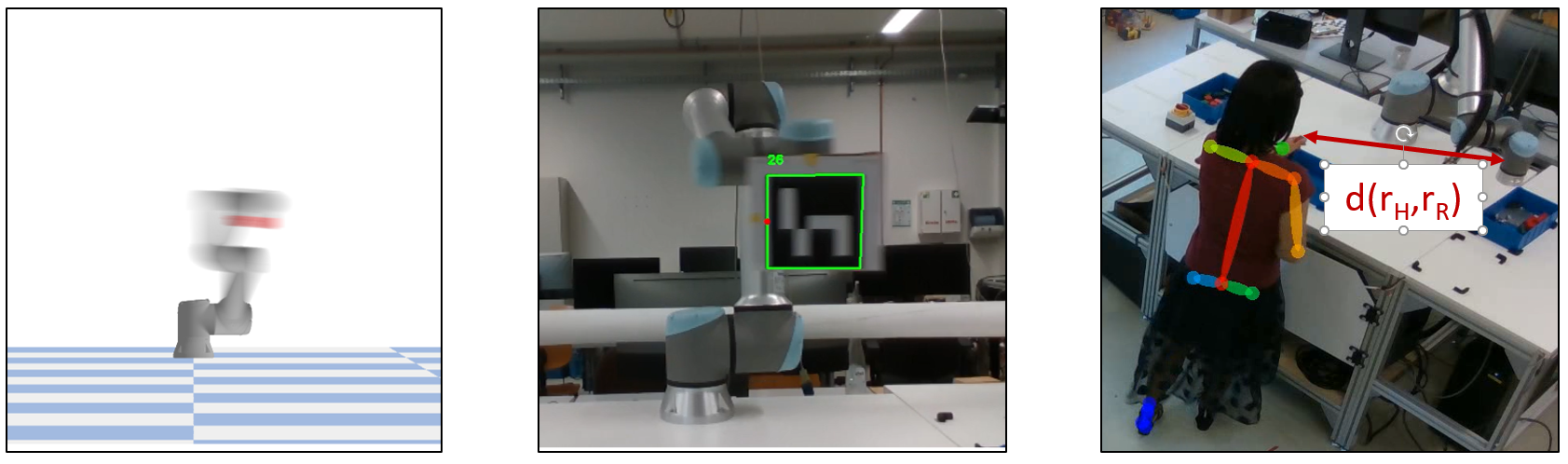}
     \caption{We consider three robot velocities in PyBullet to validate the correctness of our method. It can be seen that the uncertainty of tracking the red mask arises at high velocities. The applicability of our method to real-world environments is demonstrated by tracking the ARUCO marker at different robot velocities. Finally, we calculate the PMU for the distance $d(r_H, r_R)$ between human and robot for safety evaluation purposes.}
      \label{fig:exp_all}
   \end{figure*}

\subsection{Human Joint Position Tracking in HRC}
To use the PMU for online safety evaluation in HRC, we now apply our method to a dynamic environment where both human and robot perform movements. 
In particular, we employ OpenPose 3D \cite{Cao2019} to track the human joints online. 
One challenging issue here is that the uncertainty of OpenPose is not explicitly stated since research on estimating uncertainties of neural networks is ongoing \cite{Gawlikowski2021}. 
To overcome this, we estimate the uncertainty in the human joint tracking by leveraging the knowledge of conservation properties as explained in Section \ref{subsec_uncclass} and presented in \cite{Baek2022}. 
As elaborated in Section \ref{sec_PMULambda} and stated in Eq.~\eqref{eq:attribute_distance}, evaluating safety in HRC requires the calculation of the PMU on the distance $d$ given by 
\begin{equation}
d = \sqrt{(x_H-x_R)^2+(y_H-y_R)^2+(z_H-z_R)^2}
\end{equation}
with the coordinates $x_H$, $x_R$, $y_H$, $y_R$, $z_H$, $z_R$ in 3D-Euclidean space for human and robot, respectively. 
We assume that the robot position uncertainty is negligible as robot control algorithms are relatively accurate compared to the human joint position tracking \ie, 
$
    u_d(t) = u_{\vec{r}_H}(t).
$
Consequently, the uncertainty regarding the distance between human and robot equals the human tracking uncertainty. 

\section{Results, Evaluation and Discussion} 
To discuss the correctness of our method for the PMU quantification, we estimate the discrepancy to the ground truth uncertainty by means of Eq.~\eqref{eq:discrepancy}. 
The results are shown in Table \ref{tab:results}. 
These values correspond to the averaged discrepancy over 6000 frames in simulation and 900 frames recorded for the ARUCO marker tracking. 
Interestingly, we find that the discrepancies do not correlate linearly with the robot velocity. 
While the gap to the ground truth value increases first, slight reductions can be recognized between the second and the third velocity setting. 

\subsection{Validation in Simulation} \label{eval_sim}
To analyze whether our method delivers correct results, we compare obtained values with the ground truth uncertainty. 
Since the camera calibration was performed in world coordinates, the ground truth uncertainty can be calculated via the deviation between the actual end-effector coordinates provided by the RTDE and the measured coordinates of the red mask on the end-effector. 
For the validation, we quantify the PMU with our method by considering following uncertainty sources: the frame rate of the camera $f_p$, its resolution limit and the robot speed. 
According to our results, the PMU matches with the ground truth uncertainty to $8.05\,\%$.  
The discrepancy arises from the relatively large area of the red mask. 
For our analyses, we focused on the center of the red mask while blurr effects might have led to shifts as can be recognized in Fig.~\ref{fig:exp_all}. 
Overall, the results show that our method yields reliable uncertainty estimates in a simulation environment. 
Noting that the performance must indeed be analyzed for more complex scenarios, we argue that the suggested PMU calculation can be incorporated to enhance the risk detection performance in offline simulation tools. 

\subsection{Real-World Applicability}
In analogy to the previous experiment, the ground truth uncertainty is determined in straightforward manner by calculating the deviation between the actual robot position provided by RTDE and the 2D coordinates of the ARUCO marker. 
Our results stand in accordance to $10.8\,\%$ on average with the ground truth uncertainty. 
However, the PMU does not seem to correlate linearly with the robot velocity. 
In addition, we observe that the uncertainty arises at the turning points of the trajectories. 
\begin{table}[htb]
    \caption{Discrepancy between the ground truth uncertainty and the PMU for three robot velocities}
    \centering
    \begin{tabularx}{\columnwidth}{|c||c|c|>{\centering\arraybackslash}X|} \hline
        Experiment & $v_1$ & $v_2$ & $v_3$  \\ \hline
        Simulation & $6.30\,\%$ & $9.55\,\%$ & $8.70\,\%$ \\
        Real World (ARUCO) & $9.05\,\%$ & $10.32\,\%$ & $11.03\,\%$ \\ \hline
    \end{tabularx}
    \label{tab:results}
\end{table}

\subsection{Safety Evaluation in HRC}
Since our method has been validated by above experiments for both simulation and real-world environments, we apply it for safety evaluation in a dynamic HRC scenario. 
The uncertainty of the human joint position detection with OpenPose is obtained by leveraging the conservation measures as explained in Section \ref{subsec_uncclass}. 
To calculate the PMU with Eq.~\eqref{eq:prop_corr}, we account for the two intel RealSense cameras that were employed for the 3D human joint detection. 
Specifically, its resolution limit $u_{RealSense}\leq 0.02\,\%$ is squared for the PMU calculation. 
To accomplish the safety evaluation, we map the PMU on ISO 13849 by referring to Eq.~\eqref{eq:mapping}. 
We show the relative PMU values with respect to the length of each body segment in Table \ref{tab:safety_eval}. 
It can be seen that the detection of the thighs and the spine underlie significant uncertainties of 23.6\,\%. 
Hence, considering a length of $40\,cm$ for the thighs, a deviation of more than $8\,cm$ might occur. 
We therefore conclude that the uncertainty in the human joint detection is one main bottle neck to achieve safe HRC according to ISO 13849. 
Specifically, we find that the tracking uncertainty must be reduced by at least two orders of magnitudes to satisfy this standard. 

\begin{table}[htb]
\caption{Averaged PMU values for Body Segments}
\centering
\begin{tabularx}{\columnwidth} { 
  | >{\centering\arraybackslash}X 
  || >{\centering\arraybackslash}X | }
 \hline
 Body Segment  & Averaged PMU  \\
 \hline
   Spine & 20.2\,\% \\
   Collarbones & 4.2\,\% \\
   Thighs & 23.6\,\% \\
   Arms & 15.7\,\% \\ \hline
\end{tabularx}
\label{tab:safety_eval}
\end{table}

\subsection{Limitations}
While the PMU is one efficient method to evaluate safety, we clearly note that the validation of the correctness for more complex scenarios must be performed individually. 
Especially for real-world applications, where the ground truth uncertainty is not easily accessible, we recommend to calculate the PMU separately for different parts of the system to simplify the validation procedure. 
Apart from that, the computation time for quantifying the PMU might be one critical issue for complex scenes due to the MC algorithm.  

\section{Conclusion and Outlook}
In this paper, we presented a generalizable and data-driven method to quantify the propagated measurement uncertainty in robot systems. 
Importantly, we derived how this scalar quantity, representing the possible deviation of measured values on a $95\,\%$ confidence level, can be used to evaluate quantitative safety limits online for Human-Robot Collaboration. 
One of the unique advantages of our method is given by its flexibility as it can be applied for both, simulation and real-world environments. 
In addition to the possibility of evaluating safety during operation, our method can contribute to design safe systems by adopting it into simulation frameworks that aim to identify hazards and predict situations that violate safety limits. 
In the future, we plan to address more thorough analyses of how to enhance the computation time of our algorithm for more complex scenarios with multiple person tracking.












\bibliographystyle{IEEEtran}
\bibliography{references}



\end{document}